\documentclass[11pt]{article}

\usepackage[final]{acl}

\usepackage{times}
\usepackage{latexsym}

\usepackage[T1]{fontenc}
\usepackage[utf8]{inputenc}

\usepackage{microtype}

\usepackage{inconsolata}

\usepackage{graphicx}

\newcommand{\ie}{\textit{i.e.}}
\newcommand{\eg}{\textit{e.g.}}

\newcommand{\methodname}{\textsc{FairLocator}}

\usepackage{CJKutf8}
\usepackage{colortbl}
\usepackage{booktabs}
\usepackage{arydshln}
\usepackage{enumitem}
\usepackage{multirow}
\usepackage{subfig}
\usepackage{amsmath}

\definecolor{mygray}{RGB}{226, 226, 226}
\definecolor{myred}{RGB}{252, 142, 142}
\definecolor{mygreen}{RGB}{147, 255, 143}
\definecolor{myblue}{RGB}{144, 155, 255}
\definecolor{myyellow}{RGB}{253, 253, 143}
\definecolor{mypurple}{RGB}{255, 142, 250}

\title{AI Sees Your Location---But With A Bias Toward The Wealthy World}
\author{
Jingyuan Huang$^{2\dagger}$ \quad Jen-tse Huang$^{1\dagger}$ \quad Ziyi Liu$^1$ \quad Xiaoyuan Liu$^4$ \\
\bf Wenxuan Wang$^{3\ddagger}$ \quad Jieyu Zhao$^{1\ddagger}$ \\
$^1$University of Southern California \quad \quad $^2$University of Georgia \\
$^3$University of California, Los Angeles \quad \quad $^4$Independent Researcher \\
{\small $^{\dagger}$Equal contribution \quad \quad $^{\ddagger}$Corresponding authors}
}

\begin{document}
\maketitle

\begin{abstract}
Visual-Language Models (VLMs) have shown remarkable performance across various tasks, particularly in recognizing geographic information from images.
However, VLMs still show regional biases in this task.
To systematically evaluate these issues, we introduce a benchmark consisting of 1,200 images paired with detailed geographic metadata.
Evaluating four VLMs, we find that while these models demonstrate the ability to recognize geographic information from images, achieving up to 53.8\% accuracy in city prediction, they exhibit significant biases.
Specifically, performance is substantially higher for economically developed and densely populated regions compared to less developed (-12.5\%) and sparsely populated (-17.0\%) areas.
Moreover, regional biases of frequently over-predicting certain locations remain.
For instance, they consistently predict Sydney for images taken in Australia, shown by the low entropy scores for these countries.
The strong performance of VLMs also raises privacy concerns, particularly for users who share images online without the intent of being identified.
Our code and dataset are publicly available at \url{https://github.com/uscnlp-lime/FairLocator}.
\end{abstract}

\section{Introduction}

Visual Language Models (VLMs) have demonstrated the capability to comprehend visual content and respond to related queries~\cite{bubeck2023sparks, chow2025physbench}.
Their applications span text recognition~\cite{liu2024ocrbench, chen2025ocean}, solving mathematical problems~\cite{yang2024mathglm, peng2024multimath}, and providing medical services~\cite{azad2023foundational, buckley2023multimodal}.
Furthermore, recent research has identified their ability to infer geographic information about the location depicted in an image~\cite{wazzan2024comparing, mendes2024granular}.

However, the geographic information produced by VLMs often contains inaccuracies and significant biases~\cite{haas2024pigeon}.
These biases pose a critical issue, as they can perpetuate stereotypes about certain regions and amplify the dominance of specific areas in information dissemination~\cite{cinelli2021echo}.
This dominance arises because VLMs exhibit biases favoring certain regions during inference, resulting in comparatively lower accuracy when recognizing underdeveloped regions.
Given that VLMs are increasingly integrated into modern search engines, this imbalance strengthens users' impressions of cities that VLMs frequently or accurately identify through the mere exposure effect~\cite{zajonc1968attitudinal}, further entrenching these cities' dominance in information dissemination.

\begin{figure}[t]
    \centering
    \includegraphics[width=1.0\linewidth]{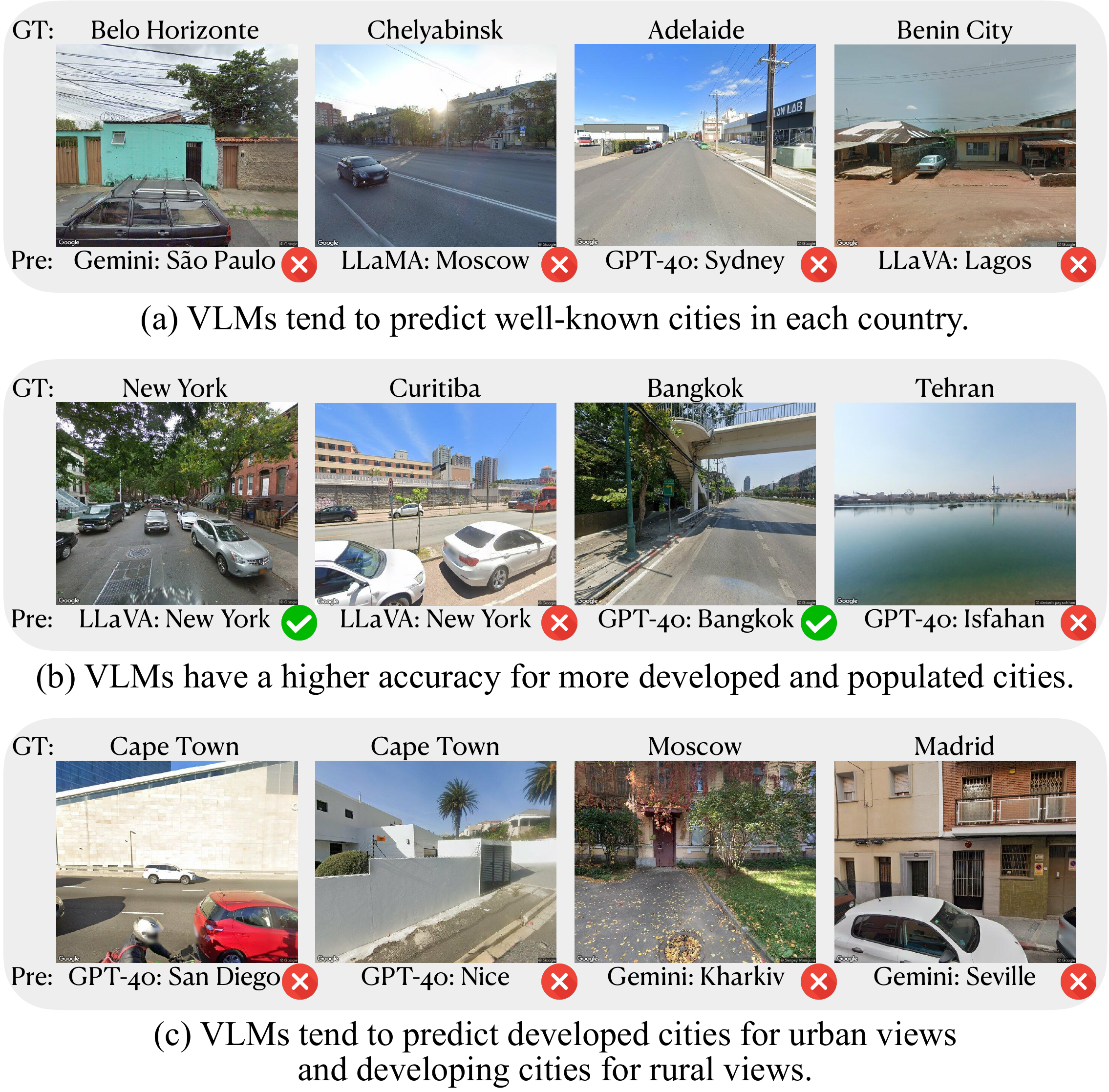}
    \caption{The three types of biases identified in this paper. ``GT'' is the ground truth while ``Pre'' represents the VLM predictions.}
    \label{fig:cover}
\end{figure}

Existing studies~\cite{liu2024image, haas2024pigeon, yang2024geolocator} have explored the ability of VLMs to recognize geographic information from images but lack a sufficient attention to bias.
Specifically, these studies fail to thoroughly analyze the biases present in VLMs' geographic information recognition.
To address this gap, we conduct a systematic investigation into the capabilities and biases of VLMs in geographic information recognition.
We categorize VLM biases in geographic information recognition into two types: (1) disparities in accuracy when identifying images from different regions and (2) systematic tendencies to predict certain regions more frequently during geographic inference.
To evaluate these biases, we develop a benchmark, {\methodname}, comprising 1,200 images from 111 cities across 43 countries, sourced from Google Street View.\footnote{\url{https://www.google.com/streetview/}}
Each image is accompanied by detailed geographic information, including country, city, and street names.
{\methodname} incorporates a benchmark to automatically query VLMs, extract responses, and align them with ground truth data using name translation and deduplication.

The dataset is divided into two subsets:
\textbf{(1) Depth}: To verify whether VLMs exhibit a tendency to predict famous cities for similar cities (\ie, cities within the same country), we select the six most populous countries from each continent and further choose ten cities from each country.
A biased model may predominantly predict well-known cities, such as Tokyo or Osaka for images of Japanese cities.
\textbf{(2) Breadth}: To explore countries with diverse cultures, populations, and development levels, we select 60 cities from a worldwide city list, ranked by population, with a maximum of two cities per country to prevent overrepresentation of highly populated nations.
Four VLMs—GPT-4o~\cite{gpt4o}, Gemini-1.5-Pro~\cite{gemini15}, LLaMA-3.2-11B-Vision~\cite{llama32}, and LLaVA-v1.6-Vicuna-13B~\cite{llava16}—are evaluated using {\methodname}.

We find that current VLMs exhibit notable biases in three key aspects:
\textbf{(1) Bias toward well-known cities}: For instance, Gemini-1.5-Pro frequently predicts São Paulo for images from Brazil. While this indicates the model's ability to recognize Brazilian features, it lacks the capacity to capture regional diversity or subtle distinctions.
\textbf{(2) Disparities in accuracy across regions}: VLMs exhibit higher performance when identifying geographic information from images of developed regions, with an average accuracy of 48.8\%, but their performance drops markedly for less developed regions, where accuracy typically falls to 41.7\%.
Similarly, the average error distance of Gemini-1.5-Pro for developed cities is 399.12 kilometers, which increases to 806.42 kilometers for developing cities.
\textbf{(3) Spurious correlations with development levels}: VLMs often associate urban or modern scenes—even from developing countries—with developed nations.
Conversely, images depicting suburban or rural views are frequently misclassified as originating from developing countries.
Our contributions in this paper are as follows:
\begin{enumerate}[leftmargin=*]
    \item We reveal, for the first time, biases in the geolocation capabilities of VLMs, which have the potential to perpetuate stereotypes among users.
    \item We develop and publish {\methodname}, a benchmark designed to facilitate future research on VLM geographical ability.
    \item We evaluate the performance of four widely-used VLMs and provide in-depth analyses to better understand their behavior.
\end{enumerate}
\section{Related Work}

\subsection{Geo-Information with AI Models}

Recent advancements in geographical information processing have leveraged Large Language Models (LLMs) and VLMs to improve geolocation tasks.
Geo-seq2seq~\cite{zhang2023geo} and \citet{hu2023geo} develop models for extracting geographical information from social media.
GPT4GEO~\cite{roberts2023gpt4geo} and \citet{bhandari2023large} explore LLMs' geographical knowledge, reasoning abilities, and spatial awareness, while GPTGeoChat~\cite{mendes2024granular}, K2~\cite{deng2024k2}, PIGEON~\cite{haas2024pigeon}, ETHAN~\cite{liu2024image}, and \citet{ramrakhiyani2025gauging} enhance models' geographical ability.
GeoLM~\cite{li2023geolm} links textual data with spatial information from geographical databases for reasoning, while GeoLLM~\cite{manvi2024geollm} integrates OpenStreetMap data to improve geospatial prediction accuracy and scalability.
GeoLocator~\cite{yang2024geolocator} and \citet{luo2025doxing} use GPT-4 and ChatGPT-o3 to infer location information from images from social media and famous landmarks, highlighting geographical privacy risks.
\citet{wazzan2024comparing} compare LLM-based search engines to traditional ones in image geolocation tasks.
Studies~\cite{shi2024assessment, zhang2024benchmarking} have also explored the use of VLMs to identify the relative positional information of objects in images, but they are not related to city-level geolocation tasks.
While these papers demonstrate significant progress in geolocation, they do not address biases in the geolocating ability of VLMs.

\subsection{Biases in AI Models}

Research has extensively documented biases in VLMs and text-to-image (T2I) models~\cite{luo2024bigbench, nakashima2023societal, wang2024new, fraser2024examining, ghosh2023person}.
Social biases in embedding spaces are also explored~\cite{brinkmann2023multidimensional, ross2021measuring}.
Few studies~\cite{zhang2022counterfactually, srinivasan2022worst, ruggeri2023multi} investigate multi-dimensional biases.
Notably, BiasDora~\cite{raj2024biasdora} and \citet{sathe2024unified} analyze biases across modalities, while VisoGender~\cite{hall2023visogender} provides datasets for pronoun resolution and retrieval tasks.
\citet{wolfe2023contrastive} reveal biases in emotional state perception and sexualized associations, and \citet{wolfe2022american} find a tendency for VLMs to associate whiteness with American identity.
\citet{wan2023biasasker}, \citet{ding2025gender}, \citet{shi2025fairgamer} and \citet{du2025faircoder} study gender and racial biases, while \citet{huang2025visbias}, \citet{wan2025male}, and \citet{huang2025fact} focus on gender biases in occupational contexts.
However, these studies do not address biases stemming from models' geolocation abilities.
\section{Data and Metrics in {\methodname}}

This section introduces how we collect data, design queries, and evaluate responses from VLMs.

\subsection{Collecting Data}

Street view images can be efficiently collected using APIs provided by mapping applications.
In this study, we utilize the Google Street View API\footnote{\url{https://developers.google.com/maps/documentation/streetview/}} (2019 Version) and address compliance with its terms of use in the Ethics Statement section.
Google ensures the blurring of personal identifiers, such as human faces and license plates, in its images.\footnote{\url{https://www.google.com/streetview/policy/}}
We begin by obtaining the central latitude and longitude coordinates of each city using the Google Geocoding API.\footnote{\url{https://developers.google.com/maps/documentation/geocoding/}}
Using these coordinates, the API retrieves images from randomly selected nearby coordinates in random angles, along with their corresponding geographical data.
For each city, a total of 10 images are collected.

\subsection{Querying VLMs}

To instruct VLMs to better perform the geolocation task, we draw inspiration from strategies frequently employed by GeoGuessr players.\footnote{\url{https://www.reddit.com/r/geoguessr/comments/9hzqlv/how_do_you_play_geoguessr/}}\footnote{\url{https://www.reddit.com/r/geoguessr/comments/9cakwx/how_to_get_better_at_geoguessr/}}
In the prompt, VLMs are required to infer geographical locations based on image details, such as house numbers, pedestrians, signage, language, and lighting.
For convenient post-processing, VLMs are required to return a response in JSON format containing five key fields: ``Analysis,'' ``Continent,'' ``Country,'' ``City,'' and ``Street.''
When encoding images as inputs for VLMs, we ensure that all EXIF (Exchangeable Image File Format) metadata—such as time, location, camera parameters, and author information—is removed, as this data could enable VLMs to infer the location easily.
Then we extract answers from outputs and ensure they are neither unknown nor invalid.
Each model is allowed up to five attempts per image; if all five attempts yield invalid results, the image is marked as a failure.
To ensure experimental reliability, each image is required to obtain three responses generated by one model.
The specific prompt used in this task is outlined below:

\begin{table}[h]
    \centering
    \label{tab-example-prompt}
    \resizebox{1.0\linewidth}{!}{
    \begin{tabular}{lp{8cm}}
    \toprule
    \rowcolor{mygray}
    \multicolumn{2}{l}{\textbf{Prompt for Geolocation Task}} \\
    \textsc{System} & Please analyze the street view step-by-step using the following criteria: (1) latitude and longitude, (2) sun position, (3) vegetation, (4) natural scenery, (5) buildings, (6) license plates, (7) road directions, (8) flags, (9) language, (10) shops, and (11) pedestrians. Provide a detailed analysis based on these features. Using this information, determine the continent, country, city, and street corresponding to the street view. \\
    \midrule
    \textsc{User} & The location names should be provided in English. Avoid special characters in your response. Please reply in JSON format using this structure: {``Analysis'': ``YourAnswer'', ``Continent'': ``YourAnswer'', ``Country'': ``YourAnswer'', ``City'': ``YourAnswer'', ``Street'': ``YourAnswer''} \\
    \bottomrule
    \end{tabular}
    }
\end{table}

\subsection{Post-Processing for Evaluation Metrics}

\begin{figure*}[t]
    \centering
    \includegraphics[width=1.0\linewidth]{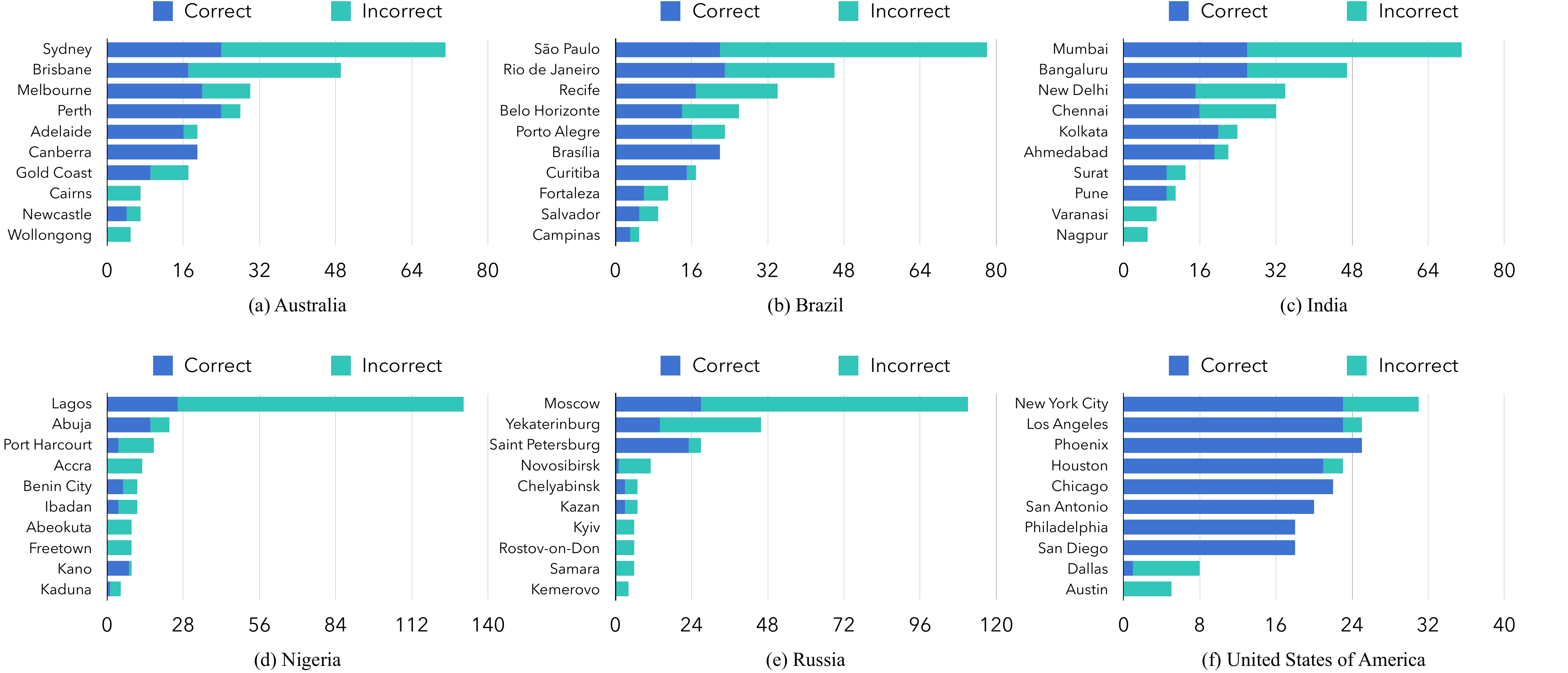}
    \caption{The most frequently predicted cities by GPT-4o across six countries. Each country includes ten cities, with ten images per city used for testing. The maximum ``Correct'' score for a city is 30, as VLMs have three attempts to predict the location. The results of other models are in \S\ref{sec:city_results} of the appendix.}
    \label{fig:city-gpt-4o}
\end{figure*}

\paragraph{Accuracy}
Since the raw text may include variations in naming or translations of the same place, we utilize GPT-4o for semantic matching in addition to exact matching for the answers.
For each image, we first attempt exact matching; if it fails, GPT-4o is employed to identify valid matches through synonyms (\eg, New York and New York City), multilingual equivalents (\eg, \begin{CJK}{UTF8}{gbsn}北京\end{CJK}, Beijing in English), and historical toponyms (\eg, Bengaluru, previously known as Bangalore).

\paragraph{Error Distance}
We use the Google Geocoding API to extract country- and city-level coordinates from VLM responses.
The geodesic distance between predicted and ground-truth coordinates is then computed.
For each image, the error distance is averaged over three independent queries.
If a prediction yields an ``unknown'' city, the error distance is set to the maximum possible on Earth—20,015 km, the distance between antipodal points.

\paragraph{Entropy}
To investigate whether VLMs exhibit bias by favoring specific cities in predictions for images from the same country, we compute the normalized entropy of the model's city-level output distribution:
$-\frac{\sum_{i=1}^{n} p_i \log_2 p_i}{\log_2 n}$,
where $p_i$ is the frequency of the $i$-th city and $n$ is the number of unique cities predicted.
This metric, based on Shannon entropy~\cite{shannon1948mathematical}, ranges from 0 to 1, with higher values indicating more uniform (\ie, less biased) predictions.
\section{Experiments}

Using {\methodname}, we focus on addressing two key research questions in this section:
(1) Do VLMs exhibit preferences for specific cities within a shared cultural background, such as within a single country (\S\ref{sec:exp-depth})?
(2) How do performance vary across global regions, considering economic, population, or cultural differences (\S\ref{sec:exp-breadth})?

\subsection{Depth Evaluation}
\label{sec:exp-depth}

The ``Depth'' subset of {\methodname} includes the most populous countries from each continent: Australia (Oceania), Brazil (South America), the United States of America (North America), Russia (Europe), and Nigeria (Africa).
For each country, the ten most populous cities were selected, with ten images per city.
Fig.~\ref{fig:city-gpt-4o} presents the cities most frequently predicted by GPT-4o, while Fig.~\ref{fig:city-gemini}, \ref{fig:city-llama}, and \ref{fig:city-llava} in \S\ref{sec:city_results} of the appendix display results from Gemini-1.5-Pro, LLaMA-3.2-11B-Vision, and LLaVA-v1.6-13B, respectively.
Notably, we exclude results from Phi-4-Multimodal~\cite{phi4mini-phi4multimodal} since it consistently outputs ``Unknown'' for all city-level queries.
We use a \texttt{temperature} of 1.0 for models except LLaVA, whose temperature is set to 0.2.
The \texttt{top\_p} is set to 1.0 for models except Gemini, who applies 0.95.

\textbf{Bias toward larger cities is observed in VLMs predictions, particularly for Brazil, Nigeria, and Russia.}
For instance, in the Nigeria test set, Lagos images constitute 10\% of the dataset, yet GPT-4o predicts ``Lagos'' 131 times, representing 43.7\% of its responses.
However, Nigerian cities such as Nnewi or Uyo (the capital of Akwa Ibom) are never predicted by GPT-4o.
Similarly, in Brazil, Gemini-1.5-Pro predicts ``S\~ao Paulo'' 181 times, accounting for 60.3\% of its predictions.
For the Russia and India test sets, Moscow and Mumbai dominate VLM predictions.
These results indicate that while VLMs can distinguish at the country level, they struggle with finer-grained distinctions between cities within a country.
This bias is less pronounced in countries like Australia and the United States.
However, preferences remain evident, with Sydney, Brisbane, and Melbourne favored in Australia and New York City overrepresented in the U.S., despite seemingly more balanced predictions.
To quantify this bias, Table~\ref{tab:entropy} shows the normalized entropy of the four models across the six countries, where scores of Nigeria and Russia are consistently lower.

\begin{table}[t]
    \centering
    \resizebox{1.0\linewidth}{!}{
    \begin{tabular}{lcccc}
        \toprule
        \bf Country & \bf GPT-4o & \bf Gemini & \bf LLaMA & \bf LLaVA \\
        \midrule
        Australia     & 0.709 & 0.745 & 0.758 & 0.781 \\
        Brazil        & 0.714 & 0.753 & 0.790 & 0.802 \\
        India         & 0.708 & 0.729 & 0.725 & 0.774 \\
        Nigeria       & \underline{0.647} & \underline{0.528} & \underline{0.643} & \underline{0.546} \\
        Russia        & 0.655 & 0.658 & 0.691 & 0.711 \\
        USA           & \textbf{0.800} & \textbf{0.803} & \textbf{0.822} & \textbf{0.808} \\
        \bottomrule
    \end{tabular}
    }
    \caption{Normalized entropy in ``Depth'' evaluation. Highest scores across countries are marked in \textbf{bold} while lowest are \underline{underlined}.}
    \label{tab:entropy}
\end{table}

\textbf{As model capabilities increase, VLMs demonstrate a greater ability to discern subtle differences between similar cities.}
Fig.~\ref{fig:city-llava} highlights the performance of the weakest model, LLaVA, which predicts S\~ao Paulo, Mumbai, Lagos, Moscow, and New York City as representative of Brazil, India, Nigeria, Russia, and the U.S., respectively.
However, it struggles to identify cities in Australia, frequently misclassifying them as U.S. cities such as New York City, Miami, San Francisco, or Los Angeles.
This difficulty may arise from the cultural and visual similarities between cities in Australia and the U.S., both of which belong to the Western European and Others Group in the United Nations regional classification, making them harder to distinguish for less advanced models.

Tables~\ref{tab:distance-depth} and~\ref{tab:accuracy-depth} (in the appendix) quantify this performance in terms of normalized error distance and accuracy, respectively.
To account for differences in land area across countries, error distances are normalized by the square root of each country's land area.
Unlike accuracy, higher error distances indicate poorer performance.
Among the evaluated models, the U.S. consistently shows the lowest error distance, whereas Nigeria has the highest.
Interestingly, Australia exhibits relatively high error, likely due to its sparse urban distribution.
GPT-4o achieves the highest accuracy among the four models, outperforming the least accurate model, LLaVA, by improving continent, country, and city-level accuracy by 65.9\%, 60.4\%, and 37.4\%, respectively.
Among the countries analyzed, VLMs most effectively recognize the U.S. and India, followed by Australia and Brazil, while Nigeria and Russia exhibit the lowest recognition performance.

\begin{table}[t]
    \centering
    \resizebox{1.0\linewidth}{!}{
    \begin{tabular}{lcccc}
        \toprule
        \bf Country & \bf GPT-4o & \bf Gemini & \bf LLaMA & \bf LLaVA \\
        \midrule
        Australia     & 0.642 & 0.488 & 1.167 & 6.661 \\
        Brazil        & 0.229 & 0.314 & 0.980 & 4.313 \\
        India         & 0.304 & 0.333 & 0.556 & 7.497 \\
        Nigeria       & \underline{0.930} & \underline{0.652} & \underline{2.039} & \underline{12.692} \\
        Russia        & 0.361 & 0.364 & 0.565 & \textbf{3.154} \\
        USA           & \textbf{0.203} & \textbf{0.130} & \textbf{0.249} & 5.856 \\
        \bottomrule
    \end{tabular}
    }
    \caption{Error distance in ``Depth'' evaluation, normalized by the square root of each country's land area. Lowest scores across countries are marked in \textbf{bold} while highest are \underline{underlined}.}
    \label{tab:distance-depth}
\end{table}

\begin{table*}[t]
    \centering
    \resizebox{1.0\linewidth}{!}{
    \begin{tabular}{llcccccccccc}
        \toprule
        \multicolumn{2}{c}{\multirow{2}{*}{\bf Models}} & \multirow{2}{*}{\bf Avg.} & \multicolumn{2}{c}{\bf Economy} & \multicolumn{2}{c}{\bf Population} & \multicolumn{5}{c}{\bf Country Group} \\
        \cmidrule(lr){4-5} \cmidrule(lr){6-7} \cmidrule(lr){8-12}
        & & & \bf Developing & \bf Developed & \bf Underpop. & \bf Populous & \bf Africa & \bf APSIDS & \bf EEG & \bf GRULAC & \bf WEOG \\
        \midrule
        \multirow{4}{*}{\rotatebox{90}{\bf GPT-4o}}
        & \bf Cont. & 90.1 & 87.1 & 96.0 & 88.1 & 93.1 & 83.1 & 91.5 & \bf 100.0 & 87.3 & 95.9 \\
        & \bf Ctry. & 81.3 & 77.8 & 88.5 & 75.3 & 90.4 & 64.4 & 85.2 & \bf 86.7 & 83.3 & 88.9 \\
        & \bf City & \bf 67.2 & \bf 64.3 & \bf 72.8 & \bf 61.1 & \bf 76.2 & 55.8 & \bf 64.2 & \bf 75.0 & \bf 73.3 & \bf 82.6 \\
        & \bf St. & \bf 3.2 & \bf 2.5 & \bf 4.5 & \bf 2.8 & \bf 3.8 & \bf 4.2 & \bf 2.1 & \bf 10.0 & \bf 2.3 & \bf 4.4 \\
        \midrule
        \multirow{4}{*}{\rotatebox{90}{\bf Gemini}}
        & \bf Cont. & \bf 95.6 & \bf 94.2 & \bf 98.2 & \bf 94.4 & \bf 97.4 & \bf 92.2 & \bf 96.2 & \bf 100.0 & \bf 93.7 & \bf 99.3 \\
        & \bf Ctry. & \bf 84.6 & \bf 81.7 & \bf 90.3 & \bf 79.4 & \bf 92.2 & \bf 73.3 & \bf 86.7 & 78.3 & \bf 85.7 & \bf 93.3 \\
        & \bf City & 61.9 & 61.7 & 62.5 & 57.5 & 68.6 & \bf62.2 & 56.5 & 66.7 & 66.3 & 71.9 \\
        & \bf St. & 2.5 & 2.0 & 3.5 & 2.2 & 2.9 & 2.5 & 1.6 & 6.7 & 0.7 & 6.3 \\
        \midrule
        \multirow{4}{*}{\rotatebox{90}{\bf LLaMA}}
        & \bf Cont. & 79.3 & 77.2 & 83.5 & 76.1 & 84.2 & 66.1 & 86.2 & 93.3 & 72.7 & 80.7 \\
        & \bf Ctry. & 60.1 & 53.6 & 73.2 & 52.9 & 71.0 & 40.8 & 65.4 & 70.0 & 57.0 & 71.1 \\
        & \bf City & 35.3 & 33.2 & 39.7 & 28.5 & 45.6 & 24.2 & 36.8 & 51.7 & 33.3 & 44.4 \\
        & \bf St. & 0.1 & 0.0 & 0.2 & 0.1 & 0.0 & 0.0 & 0.0 & 0.0 & 0.0 & 0.4 \\
        \midrule
        \multirow{4}{*}{\rotatebox{90}{\bf LLaVA}}
        & \bf Cont. & 44.4 & 40.3 & 52.7 & 39.8 & 51.4 & 17.5 & 52.6 & 95.0 & 33.3 & 57.0 \\
        & \bf Ctry. & 21.4 & 15.8 & 32.5 & 16.9 & 28.1 & 11.7 & 22.2 & 20.0 & 12.0 & 42.6 \\
        & \bf City & 11.8 & 7.7 & 20.2 & 6.9 & 19.3 & 7.2 & 11.1 & 6.7 & 6.7 & 27.0 \\
        & \bf St. & 0.0 & 0.0 & 0.0 & 0.0 & 0.0 & 0.0 & 0.0 & 0.0 & 0.0 & 0.0 \\
        \midrule
        \midrule
        \multirow{4}{*}{\rotatebox{90}{\bf Avg.}}
        & \bf Cont. & 77.3 & 74.7 & 82.6 & 74.6 & 81.5 & 64.7 & 81.6 & 97.1 & 71.8 & 83.2 \\
        & \bf Ctry. & 61.8 & 57.2 & 71.1 & 56.1 & 70.4 & 47.6 & 64.9 & 63.7 & 59.5 & 74.0 \\
        & \bf City & 44.1 & 41.7 & 48.8 & 38.5 & 52.4 & 37.4 & 42.2 & 50.0 & 44.9 & 56.5 \\
        & \bf St. & 1.4 & 1.1 & 2.0 & 1.3 & 1.7 & 1.7 & 0.9 & 4.2 & 0.8 & 2.8 \\
        \bottomrule
    \end{tabular}
    }
    \caption{Accuracy of the four models in the ``Breadth'' evaluation. ``Cont.'' represents continent, ``Ctry.'' denotes country, and ``St.'' is street. ``Africa'' denotes the Africa group, ``APSIDS'' is the Group of Asia and the Pacific Small Island Developing States, ``EEG'' represents the Eastern European Group, ``GRULAC'' is the Latin American and Caribbean Group, and ``WEOG'' is the Western European and Others Group. Best models are marked in \textbf{bold}.}
    \label{tab:accuracy-breadth}
\end{table*}

\begin{table*}[t]
    \centering
    \resizebox{1.0\linewidth}{!}{
    \begin{tabular}{llcccccccccc}
        \toprule
        \multicolumn{2}{c}{\multirow{2}{*}{\bf Models}} & \multirow{2}{*}{\bf Avg.} & \multicolumn{2}{c}{\bf Economy} & \multicolumn{2}{c}{\bf Population} & \multicolumn{5}{c}{\bf Country Group} \\
        \cmidrule(lr){4-5} \cmidrule(lr){6-7} \cmidrule(lr){8-12}
        & & & \bf Developing & \bf Developed & \bf Underpop. & \bf Populous & \bf African & \bf APSIDS & \bf EEG & \bf GRULAC & \bf WEOG \\
        \midrule
        \bf GPT-4o & \bf Ctry. & 1053.1 & 1225.8 & 707.8 & 1317.7 & 656.3 & 1787.2 & 882.7 & 158.0 & 1332.8 & 473.9 \\
        \bf Gemini & \bf Ctry. & \bf 670.7 & \bf 806.4 & \bf 399.1 & \bf 809.2 &  \bf 462.8 &  \bf 1070.7 & \bf 548.4 & \bf 243.1 & \bf 1103.4 & \bf 118.2 \\
        \bf LLaMA & \bf Ctry. & 2460.2 & 2760.4 & 1859.9 & 2787.9 & 1968.9 & 3734.0 & 1825.0 & 1004.6 & 3368.2 & 1982.5 \\
        \bf LLaVA & \bf Ctry & 10353.6 & 10388.7 & 10283.4 & 11162.4 & 9140.3 & 12419.2 & 9838.7 & 5958.0 & 10869.3 & 9547.8 \\
        \midrule
        \midrule
        \bf Avg. & \bf Ctry. & 3634.4 & 3795.3 & 3312.5 & 4019.3 & 3057.1 & 4752.7 & 3273.7 & 1840.9 & 4168.4 & 3030.6 \\
        \bottomrule
    \end{tabular}
    }
    \caption{Error distance of the four models in the ``Breadth'' evaluation. Best models are marked in \textbf{bold}.}
    \label{tab:distance-breadth}
\end{table*}

Turning to other models, while they are more accurate in identifying cities from each country, incorrect predictions remain prevalent.
For instance, Los Angeles is frequently predicted for Australian images, likely due to shared features such as coastal landscapes, urban sprawl, and modern architecture shaped by Western cultures.
Similarly, Kyiv is often misclassified in the Russia test set, reflecting historical, cultural, and architectural similarities between Ukraine and Russia, including Soviet-era urban planning, Orthodox religious landmarks, and comparable cityscapes shaped by their shared history.
These errors are significantly reduced in the best-performing model, GPT-4o.

\subsection{Breadth Evaluation}
\label{sec:exp-breadth}

The ``Breadth'' subset of {\methodname} comprises 60 cities selected based on their population rankings, starting from the highest.
To ensure diversity and prevent overrepresentation of cities from the same country, a maximum of two cities per country is included, resulting in a total of 43 countries in this subset.
This extends beyond the six countries represented in the ``Depth'' subset.
To investigate regional variations in VLM predictions, each city is further classified based on its economic status, population size, and cultural context:
\textbf{(1) Economic status} is determined using a global ranking of cities by the number of millionaires.\footnote{\url{https://www.henleyglobal.com/publications/wealthiest-cities-2024}}
The top 50 cities on this list are categorized as ``Developed'' cities, yielding 20 developed cities and 40 developing cities in the subset.
\textbf{(2) Population size} is annotated based on a global population ranking of cities.\footnote{\url{https://worldpopulationreview.com/cities}}
Cities with populations exceeding 10 million are classified as ``Populous,'' resulting in 22 populous and 38 less populous cities.
\textbf{(3) Cultural classification}: Continents are usually deemed insufficient as a standard due to the cultural diversity within them.
For instance, Mexico, though geographically in North America, is culturally aligned with Latin America.
Similarly, the U.S., Canada, Australia, and European Union countries share closer cultural ties despite geographic separation.
Therefore, the United Nations Regional Groups\footnote{\url{https://en.wikipedia.org/wiki/United_Nations_Regional_Groups}} categorization is adopted, which categorizes countries into five culturally related groups: Africa Group, APSIDA, EEG, GRULAC, and WEOG.
Table~\ref{tab:accuracy-breadth} provides the definitions of each group in its caption.

The accuracy and error distance, categorized by economic, population, and cultural groups, are separately presented in Table~\ref{tab:accuracy-breadth} and~\ref{tab:distance-breadth}.
For accuracy, the performance at city level is higher (44.1\%) compared to the ``Depth'' evaluation (25.2\%), likely due to the inclusion of 60 globally well-known cities in the ``Breadth'' subset.
Unlike the ``Depth'' evaluation, where GPT-4o performed best, the ``Breadth'' evaluation shows comparable performance between Gemini-1.5-Pro and GPT-4o.
Gemini excels at identifying continents and countries, while GPT-4o demonstrates superior performance in recognizing cities.
For error distance, Gemini outperforms all other models while LLaVA shows obviously worse performance than the other three models.
Regarding biases toward developed, populous cities and those within specific cultural groups, the key findings are as follows:

\textbf{(1) All four models consistently demonstrate lower accuracy and higher error distance in developing and less populous cities, with population exerting a greater influence on performance.}
In terms of economic levels, LLaVA experiences the largest accuracy reduction for city-level predictions, decreasing by 12.5\% when shifting from developed to developing cities.
LLaMA experiences the largest distance increase, increasing by 901.65 km when shifting from developed to developing cities.
Conversely, Gemini is least affected, with only a 0.8\% drop at the city level, although its accuracy at the country level declines by 8.6\%.
This may be due to LLaVA's consistently poor performance in both developing and developed cities, with developed cities only marginally better than developing cities.
For population, the performance drop is more obvious.
VLMs exhibit a 12.4\% to 17.1\% decrease in city-level prediction accuracy and 962.8 km increase in error distance when transitioning from more populous to less populous cities.

\textbf{(2) Accuracy and error distance vary significantly between cultural groups, with city-level accuracy and error distance differing by up to 19.1\% and 2911.9 km.}
WEOG countries achieve the highest average city-level accuracy (56.5\%), followed by EEG (50.0\%), while the Africa Group exhibits the lowest accuracy (37.4\%).
Similarly, EEG countries achieve the lowest average error distance (1841.7 km), followed by WEOG (3031.5 km), while the African Group exhibits the highest error distance (4753.6 km).
This pattern holds for most VLMs, with the exception of Gemini, where the distance order of EEG and WEOG differs, highlighting the underrepresentation of African countries in VLMs' parametric knowledge.
For accuracy, Gemini demonstrates the smallest disparity in accuracy between the Africa Group and WEOG (9.7\%), whereas GPT-4o shows the largest disparity (26.8\%).
For error distance, Gemini demonstrates the smallest disparity in error distance between the African Group and WEOG (952.94 km), whereas LLaVA shows the largest disparity (2871.34 km).

\subsection{Error Analysis with Confusion Matrix}

We computed a continent-level confusion matrix over all test predictions (Depth and Breadth) from using GPT-4o (other results are listed in \S\ref{sec:confusion}) of the appendix, allowing us to examine both near-miss and intercontinental misclassifications.
As shown in Table~\ref{tab:gpt_confusion}, the majority of predictions fall within the correct continent.
In particular, Europe (90.95\%), North America (98.07\%), and South America (92.25\%) exhibit high within-region accuracy.
While Asia and Africa show slightly higher intercontinental confusion (\eg, Asia to Europe at 12.36\%, Africa to Asia at 5.61\%), the model still generally avoids large cross-continental errors.
These results suggest that when errors occur, they often involve geographically or culturally proximate regions—reinforcing the model's partial geographic awareness even in failure cases

\begin{table}[t]
    \centering
    \resizebox{1.0\linewidth}{!}{
    \begin{tabular}{lcccccc}
    \toprule
    \textbf{Continent} & \textbf{Africa} & \textbf{Asia} & \textbf{Europe} & \textbf{NA} & \textbf{SA} & \textbf{Oceania} \\
    \midrule
    Africa & \textbf{88.48} & 5.61 & 1.36 & 1.52 & 2.88 & 0.15 \\
    Asia & 1.06 & \textbf{83.82} & 12.36 & 1.95 & 0.73 & 0.08 \\
    Europe & 0.00 & 6.43 & \textbf{90.95} & 1.67 & 0.00 & 0.95 \\
    NA & 0.00 & 0.72 & 0.00 & \textbf{98.07} & 0.72 & 0.48 \\
    SA & 0.19 & 0.58 & 1.36 & 5.62 & \textbf{92.25} & 0.00 \\
    Oceania & 0.56 & 0.00 & 2.50 & 6.39 & 2.22 & \textbf{88.31} \\
    \bottomrule
    \end{tabular}
    }
    \caption{Confusion matrix of the continent-level results from GPT-4o. NA and SA: North and South America.}
    \label{tab:gpt_confusion}
\end{table}

\subsection{User Study}

To demonstrate the difficulty of recognizing images in {\methodname}, we conduct a user study using a randomly sampled subset of 1,200 images.
From this subset, 100 images are selected and organized into ten questionnaires, each containing ten images.
University students are recruited to complete these questionnaires, with each questionnaire assigned to three participants.
Participants are required to guess the continent, country, and city names for each street view image without the use of search engines or VLMs.
An example questionnaire is provided in Fig.~\ref{fig:questionnaire} in the appendix.
Table~\ref{tab:human} reports human accuracy, \textbf{revealing significantly lower performance compared to VLMs.}
Specifically, the best-performing model, Gemini-1.5-Pro, outperformed humans by 59.6\%, 74.2\%, and 62.6\% in continent, country, and city-level predictions, respectively.
Most human participants report having no familiarity with the images and indicate that their responses are purely guesswork.
These findings highlight the superiority of VLMs' parametric knowledge over human capabilities, enabling common users to easily identify geolocation, thereby increasing the risk of privacy exposure.

\begin{table}[t]
    \centering
    \resizebox{0.8\linewidth}{!}{
    \begin{tabular}{lccc}
        \toprule
        \bf Model & \bf Continent & \bf Country & \bf City \\
        \midrule
        \bf GPT-4o & 86.0 & 74.0 & 63.3 \\
        \bf Gemini & \bf 93.3 & \bf 83.7 & \bf 64.3 \\
        \bf LLaMA & 76.7 & 59.0 & 32.3 \\
        \bf LLaVA & 45.0 & 21.0  & 11.0 \\
        \hdashline
        \bf Human & 33.7 & 9.5 & 1.7 \\
        \bottomrule
    \end{tabular}
    }
    \caption{VLMs and human performance on a small subset (100 images) of {\methodname}. Highest scores are marked in \textbf{bold}.}
    \label{tab:human}
\end{table}
\section{Further Analyses}

\subsection{Is There Data Leakage?}

\paragraph{Newer Version of Images}

Given the exceptional performance of VLMs, one might hypothesize that Google Street View images are included in their training data, leading to potential memorization of answers.
To investigate this, we supplement the 2019 version of Google Street View images used in the main experiments with a newer version from 2024 and an older version from 2014.
The 2024 images are not included in the training data of GPT-4o and Gemini-1.5-Pro, as their release dates postdate those of the models.
The inclusion of 2014 images aims to introduce more varied street views.
Given the limited availability of some versions in certain regions, we select three U.S. cities, \ie, Denver, Las Vegas, and New York.
Results show that, in terms of city-level accuracy, GPT-4o achieved an accuracy of 79.1\% for the 2014 images, 89.1\% for the 2019 images, and 86.7\% for the 2024 images.
In contrast, Gemini attained accuracies of 79.2\% for the 2014 images, 80.0\% for the 2019 images, and 78.3\% for the 2024 images.
Notably, we observe substantial changes in buildings at three Las Vegas locations between 2014 and 2019, on which model predictions are inaccurate for 2014 imagery but accurate for 2019 and 2024.
This pattern indicates that VLMs may depend on features that change over time, which is influenced by their training data.

\paragraph{Identifying User-Uploaded Images}

In addition to utilizing the latest version of Google Street View images, we incorporate images captured by the authors, ensuring that none have previously been published online.\footnote{All image providers (authors) have granted consent for the use of these images in this research and their publication in an open repository.}
The data include six cities worldwide: Bangkok, Chicago, Los Angeles, Mexico City, Shanghai, and Sydney, with 10 images collected per city.
We evaluate the accuracy of VLMs using these user-provided images in comparison with Google Street View images from the same cities.
The results, presented in Table~\ref{tab:user-photo}, indicate that VLM achieves higher accuracy on user-provided images, particularly for those from Shanghai.
This may be attributed to the broader field of view and richer contextual information in user-provided images compared to Google Street View.
This finding strengthens the privacy concern, as VLMs could be used to identify locational information from user-uploaded images on the Internet.

\begin{table}[t]
    \centering
    \resizebox{1.0\linewidth}{!}{
    \begin{tabular}{lccc}
        \toprule
        \textbf{Country Group} & \textbf{Continent} & \textbf{Country} & \textbf{City} \\
        \midrule
        Latin American and Caribbean Group & 1   & 1   & 0.8 \\
        African Group                      & 0.9 & 0.3 & 0.3 \\
        Western European and Others Group  & 0.8 & 0.8 & 0.5 \\
        Eastern European Group             & 1   & 1   & 0.7 \\
        Asia and the Pacific Group         & 1   & 1   & 0.9 \\
        \midrule
        Average                            & 0.94 & 0.82 & 0.64 \\
        \bottomrule
    \end{tabular}
    }
    \caption{Accuracy of GPT-4o on Google Street View images of landmarks.}
    \label{tab:landmarks}
\end{table}

\paragraph{Identifying Landmarks}

We further test VLMs' geolocation capabilities with landmark-rich images depicting heritage sites, that have a higher chance to be included in training data.
To this end, we collect 50 images of globally recognized heritage sites, randomly selected from the UNESCO World Heritage List\footnote{\url{https://whc.unesco.org/en/list/}} across the five UN regional groups (10 images per group from Google Street View).
GPT-4o results are summarized in Table~\ref{tab:landmarks}.
Interestingly, while continent- and country-level accuracy is higher than daily scenes, the city-level accuracy is not consistently better than in our main experiments.
This may be attributed to the fact that many heritage sites are located in sparsely populated or rural areas, which VLMs often misclassify at the city level—similar to the biases we find in Table~\ref{tab:accuracy-breadth}.

\begin{table}[t]
    \centering
    \resizebox{1.0\linewidth}{!}{
    \begin{tabular}{lcccccc}
        \toprule
        \bf Data & \bf Bangkok & \bf Chicago & \bf LA & \bf MC & \bf Shanghai & \bf Sydney \\
        \midrule
        \multicolumn{7}{c}{\bf GPT-4o} \\
        \hdashline
        Google & 63.3 & 73.3 & 76.7 & 73.3 & 36.7 & 90.0 \\
        User & 100.0 & 100.0 & 90.7 & 66.7 & 93.3 & 76.7 \\
        \midrule
        \multicolumn{7}{c}{\bf Gemini-1.5-Pro} \\
        \hdashline
        Google & 83.3 & 93.3 & 60.0 & 80.0 & 23.3 & 73.3 \\
        User & 100.0 & 100.0 & 70.7 & 47.6 & 70.0 & 73.3 \\
        \bottomrule
    \end{tabular}
    }
    \caption{City-level accuracy of GPT-4o and Gemini on Google Street View images and user-uploaded images. ``LA'' is Los Angeles while ``MC'' is Mexico City.}
    \label{tab:user-photo}
\end{table}

\begin{table}[t]
    \centering
    \resizebox{1.0\linewidth}{!}{
    \begin{tabular}{lcccccc}
        \toprule
        \bf Model & \bf Bangkok & \bf Joburg & \bf Lima & \bf London & \bf NYC & \bf Sydney \\
        \midrule
        \bf GPT-4o & 90.0 & 56.7 & 96.7 & 86.7 & 100.0 & 100.0 \\
        \bf Gemini & 73.3 & 66.7 & 90.0 & 96.7 & 100.0 & 76.7 \\
        \bottomrule
    \end{tabular}
    }
    \caption{City-level accuracy of GPT-4o and Gemini on the Chinatown views. ``NYC'' is New York City. ``Joburg'' is Johannesburg.}
    \label{tab:chinatown}
\end{table}

\subsection{Is There Spurious Correlation?}

\paragraph{Specific Features}

Another hypothesis posits that VLMs may exploit superficial correlations in images to infer locations.
To examine the relationship between distinctive features and ground truths, we focus on Chinatowns across different cities, which share common visual elements such as Chinese characters and cultural decorations (\eg, red lanterns and Fai Chun).
For this experiment, one Chinatown is selected from each continent, with ten images sampled from each: Bangkok, Johannesburg, Lima, London, New York, and Sydney, all featuring established Chinatowns with significant Chinese communities.
Results from GPT-4o and Gemini-1.5-Pro, summarized in Table~\ref{tab:chinatown}, demonstrate strong performance by VLMs in identifying these Chinatown scenes.
This finding suggests that VLMs do not exclusively rely on obvious cues linking images to China but also leverage other nuanced features.

\paragraph{Style of City Views}

We further examine how the overall style of images influences predictions.
Specifically, we investigate whether VLMs exhibit biases, such as favoring developed cities for urban, modern street scenes and developing cities for rural, undeveloped environments.
For instance, as shown in Fig.~\ref{fig:cover}(c), GPT-4o predicts urban scenes from Cape Town, South Africa, as San Diego, USA, and Nice, France.
Conversely, for more rural images, Gemini-1.5-Pro misidentifies Moscow, Russia, as Kharkiv, Ukraine, and Madrid, Spain, as Seville, Spain.
Similarly, LLaMA demonstrates comparable errors: a clean, organized street scene from Bras\'ilia, Brazil, is predicted as Sydney, Australia, and a high-rise cityscape from Krasnoyarsk, Russia, is identified as New York, USA.
These findings reveal potential regional biases in VLMs when interpreting urban environments.
\section{Conclusion}

This study identifies three types of biases in VLM in geolocation tasks using {\methodname}, a benchmark comprising 1,200 images sourced globally from Google Street View.
Evaluation in two aspects---``Depth,'' covering six countries and 60 cities, and ``Breadth,'' spanning 43 countries and 60 cities---reveal two core takeaways:
(1) VLM predictions exhibit a bias toward larger cities, particularly in Brazil, Nigeria, and Russia.
The entropy reaches 0.82 in the U.S., while dropping to 0.54 in Brazil.
(2) Metrics vary notably across regions, with city-level accuracy differing by up to 19.1\% and error distance differing by up to 2911.9 km.
While VLMs demonstrate the capability to identify locations, this raises privacy concerns, particularly regarding the potential exposure of personal geographical information in regions where models perform more accurately.

\section*{Limitations}

This study has several limitations.
(1) It does not investigate the underlying causes of biases in geographical information recognition.
We hypothesize that these biases arise from imbalanced training datasets, where biased data contribute to the VLM's performance disparities.
To test this hypothesis, we propose conducting comparative experiments using models trained on different datasets.
Specifically, future research could compare the performance of VLMs trained in China and the United States in recognizing cities within China, providing deeper insights into whether dataset imbalance is a primary factor.
(2) The evaluation does not include all countries globally.
While we acknowledge the importance of every country, budget constraints limited our evaluation to 111 cities across 43 countries.
To mitigate this limitation, we selected countries from diverse regions, cultures, and development levels to ensure broad coverage.
Future studies can extend the evaluation by leveraging the workflow outlined in this paper.

\section*{Ethics Statements}

\subsection*{License of Google Street View Images}
\label{sec:license}

In this section, we detail how our work adheres to the Google Street View terms of use.\footnote{\url{https://about.google/brand-resource-center/products-and-services/geo-guidelines}}
The terms impose four key restrictions, addressed as follows:
(1) ``Creating data from Street View images, such as digitizing or tracing information from the imagery.''
Our work does not store or release specific Street View images.
Instead, we report aggregated statistics derived from the collected images, with a few example images included solely for illustrative purposes in this paper.
(2) ``Using applications to analyze and extract information from the Street View imagery.''
We do not employ external applications for analysis.
Instead, we rely on algorithmic methods for visual understanding of the Street View images.
(3) ``Downloading Street View images to use separately from Google services (such as an offline copy).''
Our work utilizes images directly via the Street View API and does not distribute the images as a dataset.
Instead, we release only the geographic coordinates, requiring future users to access the same images through the Street View API.
(4) ``Merging or stitching together multiple Street View images into a larger image.''
We do not merge or stitch Street View images in any form.
By adhering to these restrictions, we ensure compliance with Google’s terms of use for Street View, consistent with prior research practices~\cite{fan2023urban, gebru2017using, ki2021analyzing}.

\subsection*{Privacy Issues}

Our experimental results show that VLMs achieve higher accuracy in popular cities, suggesting that privacy concerns may be more pronounced in densely populated areas.
However, VLMs also significantly outperform human-level recognition in less populated regions, indicating that privacy risks are not confined to major urban centers.
Notably, VLMs are more effective at recognizing information from user-uploaded images than from Google Street View, even after we removed metadata—highlighting the potential privacy implications of public image sharing.
These findings underscore the broader concern that VLMs could be misused to infer individuals' locations from publicly posted images.
While our research aims to identify and highlight this risk in an academic and ethical context, we strongly oppose any malicious use of this technology.
By raising awareness, we hope to foster responsible discussion and encourage the development of safeguards that prevent unethical applications.

\subsection*{The Use of Large Language Models}

LLMs were employed in a limited capacity for writing optimization.
Specifically, the authors provided their own draft text to the LLM, which in turn suggested improvements such as corrections of grammatical errors, clearer phrasing, and removal of non-academic expressions.
LLMs were also used to inspire possible titles for the paper.
While the system provided suggestions, the final title was decided and refined by the authors and is not directly taken from any single LLM output.
In addition, LLMs were used as coding assistants during the implementation phase.
They provided code completion and debugging suggestions, but all final implementations, experimental design, and validation were carried out and verified by the authors.
Importantly, LLMs were \textbf{NOT} used for generating research ideas, designing experiments, or searching and reviewing related work.
All conceptual contributions and experimental designs were fully conceived and executed by the authors.

% \section*{Acknowledgments}

\bibliography{reference, model}

\onecolumn
\appendix

\section{More Results for the Depth Evaluation}

\subsection{Accuracy of Each Level}

\begin{table}[h]
    \centering
    % \resizebox{1.0\linewidth}{!}{
    \begin{tabular}{llccccccc}
        \toprule
        \multicolumn{2}{c}{\bf Models} & \bf Avg. & \bf Australia & \bf Brazil & \bf India & \bf Nigeria & \bf Russia & \bf USA \\
        \midrule
        \multirow{4}{*}{\rotatebox{90}{\bf GPT-4o}} & \bf Cont. & \bf 94.4 & 88.3 & 96.7 & \bf 99.3 & 95.0 & \bf 88.7 & 98.3 \\
        & \bf Ctry. & \bf 90.7 & 88.0 & 94.7 & \bf 97.0 & \bf 81.3 & \bf 86.0 & 97.3 \\
        & \bf City & \bf 40.4 & 45.0 & \bf 47.7 & 47.0 & \bf 22.0 & \bf 23.7 & 57.0  \\
        & \bf St. & \bf 0.6 & \bf 2.7 & \bf 0.3 & \bf 0.3 & 0.0 & \bf 0.3 & 0.0 \\
        \midrule
        \multirow{4}{*}{\rotatebox{90}{\bf Gemini}} & \bf Cont. & \bf 94.4 & \bf 91.0 & \bf 98.7 & 97.7 & \bf 98.0 & 81.0 & \bf 100.0 \\
        & \bf Ctry. & 86.2 & \bf 91.0 & \bf 96.0 & 92.3 & 77.7 & 60.3 & \bf 100.0 \\
        & \bf City & 35.4 & \bf 54.3 & 21.0 & \bf 49.3 & 14.7 & 15.3 & \bf 57.7 \\
        & \bf St. & 0.4 & 1.7 & 0.0 & 0.3 & 0.0 & 0.0 & \bf 0.3 \\
        \midrule
        \multirow{4}{*}{\rotatebox{90}{\bf LLaMA}} & \bf Cont. & 86.1 & 79.3 & 77.7 & 95.0 & 83.3 & 83.3 & 98.0 \\
        & \bf Ctry. & 75.4 & 77.7 & 71.0 & 93.3 & 38.3 & 76.7 & 95.3 \\
        & \bf City & 21.8 & 24.3 & 9.0 & 37.3 & 3.0 & 14.3 & 43.0 \\
        & \bf St. & 0.2 & 1.0 & 0.0 & 0.0 & 0.0 & 0.0 & 0.0 \\
        \midrule
        \multirow{4}{*}{\rotatebox{90}{\bf LLaVA}} & \bf Cont. & 34.0 & 3.3 & 38.7 & 39.0 & 39.0 & 32.7 & 51.3 \\
        & \bf Ctry. & 24.8 & 3.3 & 19.0 & 35.0 & 30.3 & 12.0 & 49.0 \\
        & \bf City & 3.0 & 0.7 & 1.3 & 5.0 & 3.0 & 1.7 & 6.3 \\
        & \bf St. & 0.0 & 0.0 & 0.0 & 0.0 & 0.0 & 0.0 & 0.0 \\
        \midrule
        \midrule
        \multirow{4}{*}{\rotatebox{90}{\bf Avg.}} & \bf Cont. & 77.2 & 65.5 & 77.9 & 82.8 & 78.8 & 71.4 & 86.9 \\
        & \bf Ctry. & 69.3 & 65.0 & 70.2 & 79.4 & 56.9 & 58.8 & 85.4  \\
        & \bf City & 25.2 & 31.1 & 19.7 & 34.7 & 10.7 & 13.8 & 41.0 \\
        & \bf St. &  0.3 & 1.3 & 0.1 & 0.2 & 0.0 & 0.1 & 0.1\\
        \bottomrule
    \end{tabular}
    % }
    \caption{Accuracy of the four models in the ``Depth'' evaluation across the six countries. ``Cont.'' represents continent, ``Ctry.'' denotes country, and ``St.'' is street. Highest scores are marked in \textbf{bold}.}
    \label{tab:accuracy-depth}
\end{table}

\clearpage

\subsection{City Predictions from Other VLMs}
\label{sec:city_results}

\begin{figure*}[h!]
    \centering
    \includegraphics[width=0.95\linewidth]{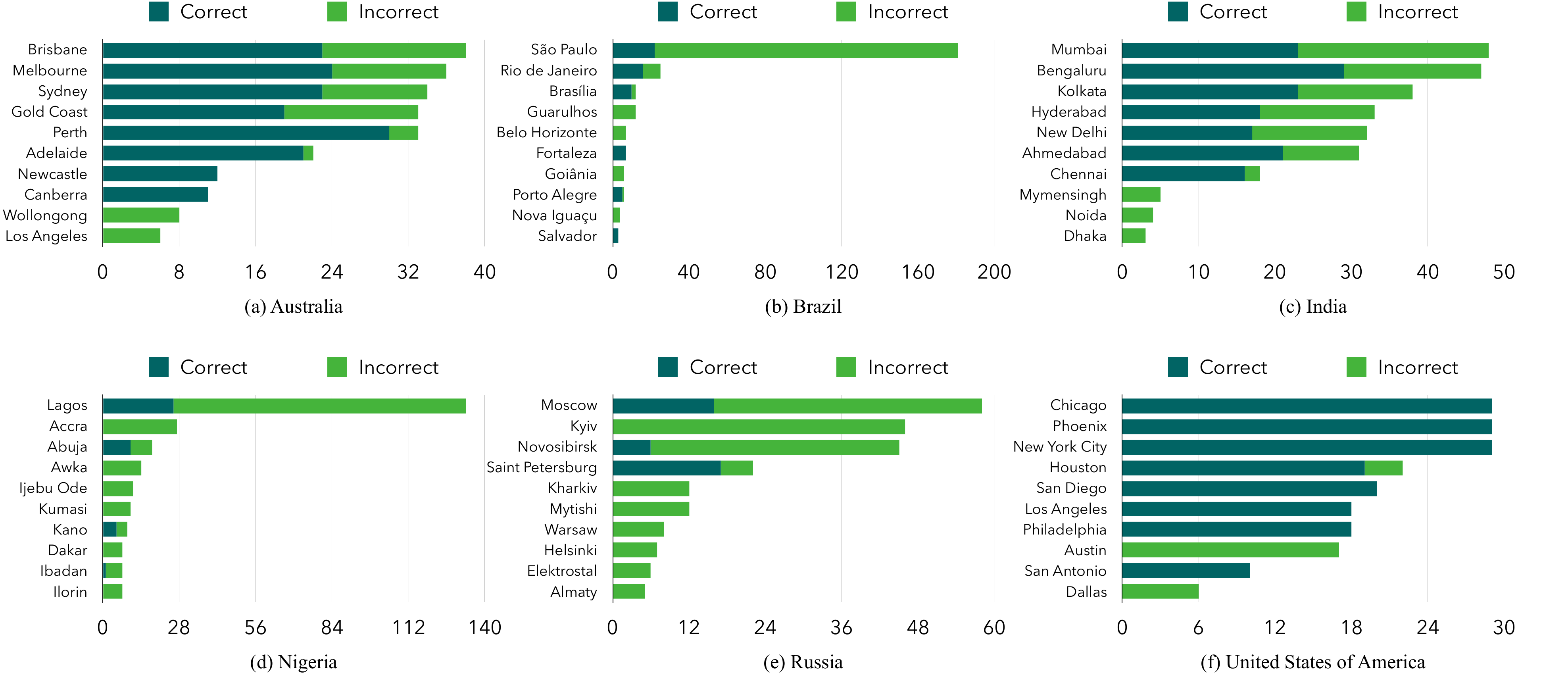}
    \caption{The most frequently predicted cities by Gemini-1.5-Pro across six countries.}
    \label{fig:city-gemini}
\end{figure*}

\begin{figure*}[h!]
    \centering
    \includegraphics[width=0.95\linewidth]{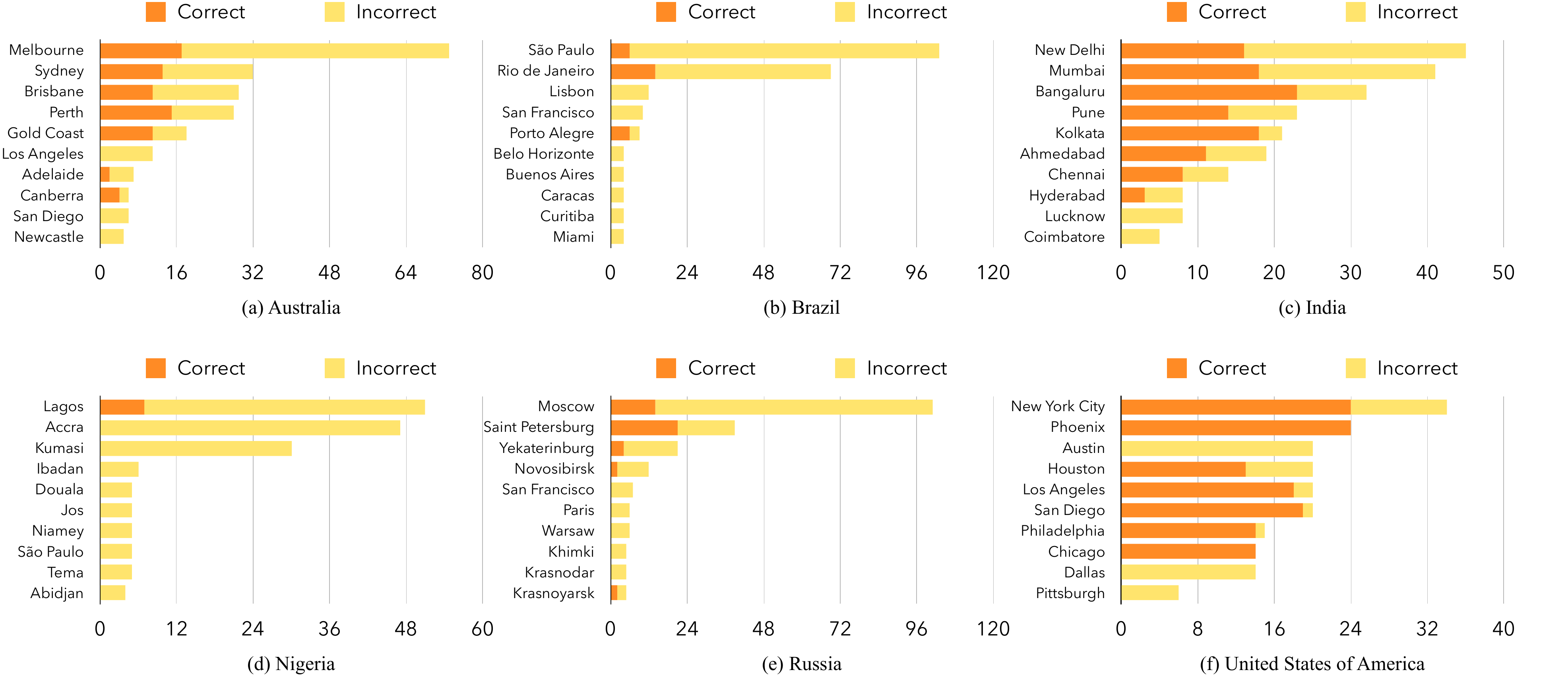}
    \caption{The most frequently predicted cities by LLaMA-3.2-11B-Vision across six countries.}
    \label{fig:city-llama}
\end{figure*}

\begin{figure*}[h!]
    \centering
    \includegraphics[width=0.95\linewidth]{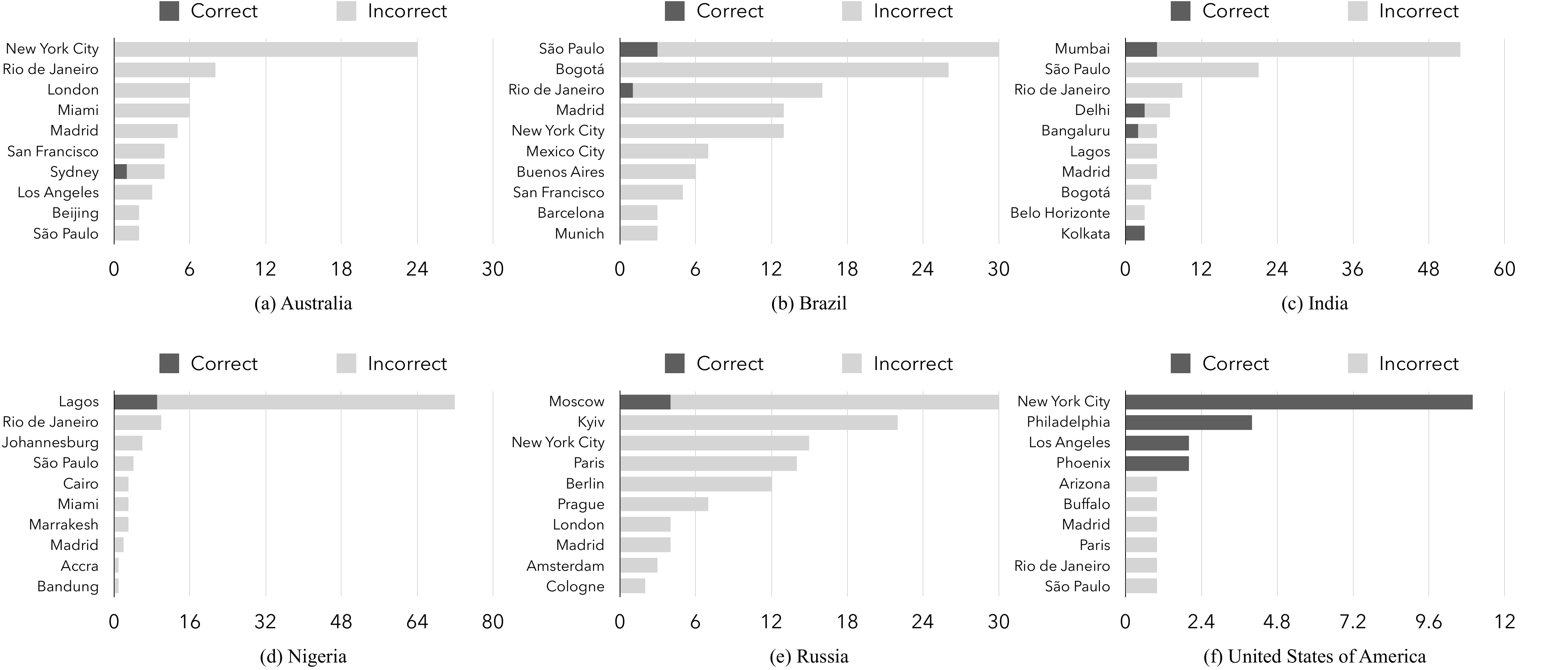}
    \caption{The most frequently predicted cities by LLaVA-V1.6-Vicuna-13B across six countries.}
    \label{fig:city-llava}
\end{figure*}

\clearpage

\subsection{Continent-Level Confusion Matrix from Other VLMs}
\label{sec:confusion}

\begin{table}[h!]
    \centering
    % \resizebox{1.0\linewidth}{!}{
    \begin{tabular}{lcccccc}
    \toprule
    \textbf{Continent} & \textbf{Africa} & \textbf{Asia} & \textbf{Europe} & \textbf{NA} & \textbf{SA} & \textbf{Oceania} \\
    \midrule
    Africa & \textbf{98.00} & 1.00 & 0.00 & 0.00 & 1.00 & 0.00 \\
    Asia & 1.56 & \textbf{66.44} & 32.00 & 0.00 & 0.00 & 0.00 \\
    Europe & 0.00 & 0.67 & \textbf{97.34} & 2.00 & 0.00 & 0.00 \\
    NA & 0.00 & 0.00 & 0.00 & \textbf{100.00} & 0.00 & 0.00 \\
    SA & 0.00 & 0.00 & 0.00 & 1.33 & \textbf{98.67} & 0.00 \\
    Oceania & 0.00 & 1.00 & 3.67 & 4.33 & 0.00 & \textbf{91.00} \\
    \bottomrule
    \end{tabular}
    % }
    \caption{Confusion matrix of the continent-level results from Gemini.}
    \label{tab:gemini_confusion}
\end{table}

\begin{table}[h!]
    \centering
    % \resizebox{1.0\linewidth}{!}{
    \begin{tabular}{lcccccc}
    \toprule
    \textbf{Continent} & \textbf{Africa} & \textbf{Asia} & \textbf{Europe} & \textbf{NA} & \textbf{SA} & \textbf{Oceania} \\
    \midrule
    Africa & \textbf{83.67} & 11.33 & 0.00 & 1.67 & 3.33 & 0.00 \\
    Asia & 3.11 & \textbf{76.22} & 15.56 & 4.00 & 0.44 & 0.22 \\
    Europe & 0.00 & 35.33 & \textbf{57.34} & 4.67 & 0.67 & 2.00 \\
    NA & 0.33 & 0.00 & 1.33 & \textbf{98.01} & 0.33 & 0.00 \\
    SA & 2.67 & 1.00 & 6.33 & 14.33 & \textbf{73.00} & 1.33 \\
    Oceania & 0.67 & 5.00 & 2.33 & 14.67 & 0.00 & \textbf{75.67} \\
    \bottomrule
    \end{tabular}
    % }
    \caption{Confusion matrix of the continent-level results from LLaMA-3.2-11B-Vision.}
    \label{tab:llama_confusion}
\end{table}

\begin{table}[h!]
    \centering
    % \resizebox{1.0\linewidth}{!}{
    \begin{tabular}{lcccccc}
    \toprule
    \textbf{Continent} & \textbf{Africa} & \textbf{Asia} & \textbf{Europe} & \textbf{NA} & \textbf{SA} & \textbf{Oceania} \\
    \midrule
    Africa & \textbf{44.67} & 3.67 & 1.00 & 3.33 & 8.67 & 0.00 \\
    Asia & 1.78 & \textbf{26.22} & 20.22 & 5.78 & 12.89 & 0.00 \\
    Europe & 0.00 & 1.33 & \textbf{64.67} & 8.00 & 0.67 & 0.00 \\
    NA & 0.67 & 0.00 & 1.00 & \textbf{50.00} & 1.00 & 0.00 \\
    SA & 0.67 & 0.67 & 12.00 & 13.33 & \textbf{38.67} & 0.00 \\
    Oceania & 0.00 & 2.33 & 6.00 & \textbf{33.67} & 4.00 & 3.00 \\
    \bottomrule
    \end{tabular}
    % }
    \caption{Confusion matrix of the continent-level results from LLaVA-V1.6-Vicuna-13B.}
    \label{tab:llava_confusion}
\end{table}

\clearpage

\subsection{Distance-Based Scores}

\begin{table}[h!]
\centering
\begin{tabular}{lcccc}
\toprule
\textbf{Country} & \textbf{Total Score} & $d <= 100km$ & $100km <d <= 1000km$ & $d > 100km$ \\
\midrule
Australia     & 415 & 174 & 67  & 59  \\
Brazil        & 407 & 159 & 89  & 52  \\
India         & 379 & 147 & 85  & 68  \\
Nigeria       & 338 & 78  & 182 & 40  \\
Russia        & 236 & 71  & 94  & 135 \\
United States & 425 & 185 & 55  & 60  \\
\bottomrule
\end{tabular}
\caption{GPT-4o scores. We define a distance-based scoring scheme as follows: Score 2: Prediction within 100 km of ground truth Score 1: Prediction within 100–1000 km Score 0: Prediction beyond 1000 km.}
\label{tab:gpt_scores}
\end{table}

\begin{table}[h!]
\centering
\begin{tabular}{lcccc}
\toprule
\textbf{Country} & \textbf{Total Score} & $d \le 100km$ & $100km < d \le 1000km$ & $d > 100km$ \\
\midrule
Australia     & 455  & 190  & 75   & 35  \\
Brazil        & 300  & 85   & 130  & 85  \\
India         & 396  & 156  & 84   & 60  \\
Nigeria       & 325  & 53   & 219  & 28  \\
Russia        & 158  & 46   & 66   & 188 \\
United States & 425  & 184  & 57   & 59  \\
\bottomrule
\end{tabular}
\caption{Gemini-1.5-Pro scores. We define a distance-based scoring scheme as follows: Score 2: Prediction within 100 km of ground truth Score 1: Prediction within 100–1000 km Score 0: Prediction beyond 1000 km.}
\label{tab:gemini_scores}
\end{table}

\begin{table}[h!]
\centering
\begin{tabular}{lcccc}
\toprule
\textbf{Country} & \textbf{Total Score} & $d \le 100km$ & $100km < d \le 1000km$ & $d > 100km$ \\
\midrule
Australia     & 297 & 100 & 97  & 103 \\
Brazil        & 185 & 41  & 103 & 156 \\
India         & 336 & 116 & 104 & 80  \\
Nigeria       & 225 & 15  & 195 & 90  \\
Russia        & 176 & 48  & 80  & 172 \\
United States & 359 & 140 & 79  & 81  \\
\bottomrule
\end{tabular}
\caption{LLaMA-3.2-11B-Vision scores. We define a distance-based scoring scheme as follows: Score 2: Prediction within 100 km of ground truth Score 1: Prediction within 100–1000 km Score 0: Prediction beyond 1000 km.}
\label{tab:llama_scores}
\end{table}

\begin{table}[h!]
\centering
\begin{tabular}{lcccc}
\toprule
\textbf{Country} & \textbf{Total Score} & $d \le 100km$ & $100km < d \le 1000km$ & $d > 100km$ \\
\midrule
Australia      & 6    & 2    & 2    & 296 \\
Brazil         & 42   & 9    & 24   & 267 \\
India          & 63   & 16   & 31   & 253 \\
Nigeria        & 92   & 9    & 74   & 217 \\
Russia         & 26   & 5    & 16   & 279 \\
United States  & 41   & 19   & 3    & 278 \\
\bottomrule
\end{tabular}
\caption{LLaVA-V1.6-Vicuna-13B scores. We define a distance-based scoring scheme as follows: Score 2: Prediction within 100 km of ground truth Score 1: Prediction within 100–1000 km Score 0: Prediction beyond 1000 km.}
\label{tab:llava_scores}
\end{table}

\twocolumn

\section{Discussions}

\subsection{Is Ten Pictures Per City Enough?}

To assess whether ten images per city are sufficient to support our conclusions, we conduct a new set of experiments using the Gemini-1.5-Pro.
Each city is represented by 20 images, with each image queried once to predict its geographical location.
To evaluate the impact of sample size reduction, we randomly select 10 images from the original 20 and compare the model’s performance to that obtained using the full set.
With 20 images per city, the model achieves a city-level accuracy of 63.0\%. Using 10 images yields an accuracy of 64.8\%, a marginal increase of 1.8 percentage points.
A per-city analysis shows that in 91.7\% of cities, the accuracy difference between the two settings is within 10\%.
Given that each image contributes 5\% to the city-level accuracy in the 20-image setting, this variation is minimal.
We also examine the stability of the model's performance.
When using 20 images per city, the mean standard deviation of city-level accuracy across cities is 0.406; with 10 images, it is 0.370—a relative difference of just 8.9\%.
This small change suggests that reducing the sample size has a negligible effect on performance variability.
Overall, the results indicate that using 10 images per city yields comparable accuracy and stability to using 20, supporting the sufficiency of smaller sample sizes for robust city-level evaluation.

\subsection{Rural vs. Urban}

To assess performance differences between urban and rural environments, we conduct a supplementary experiment involving five rural U.S. locations: Woodstock, Vermont; Smicksburg, Pennsylvania; Galena, Illinois; Barboursville, Virginia; and Blue Ridge, Georgia.
For each location, we select 10 images and query the Gemini-1.5-Pro once per image to evaluate geolocation accuracy.
The model achieves 100\% accuracy at the continent and country levels but only 3\% at the city level across these rural areas.
For comparison, we evaluate the model on 10 U.S. cities, again using 10 images per city and one query per image.
In urban settings, the model maintains 100\% accuracy at the continent and country levels and achieves 57.7\% accuracy at the city level.
These results reveal a substantial drop in city-level accuracy for rural areas, indicating that the model performs more reliably in urban regions and struggles with sparsely populated, less visually distinctive environments.
This observation reinforces our overall conclusion that geolocation accuracy improves with population density and urban visual features.

\begin{table}[t]
    \centering
    \begin{tabular}{lc}
    \toprule
    \textbf{Country} & \textbf{Accuracy} \\
    \midrule
    Australia & 14.7 \\
    Brazil & 14.0 \\
    India & 12.0 \\
    Nigeria & 14.0 \\
    Russia & 18.0 \\
    United States & 34.0 \\
    \hline
    Average & 17.8 \\
    Random Baseline & 0.9 \\
    \bottomrule
    \end{tabular}
    \caption{Accuracy of CLIP (ViT-B/32).}
    \label{tab:clip}
\end{table}

\subsection{Zero-Shot CLIP}

We have conduct an experiment using zero-shot CLIP (ViT-B/32)~\cite{clip}.
Since CLIP does not support instruction-following or structured prompting like VLMs, we adopt a retrieval-style setup.
Specifically, we pair each test image in our Depth dataset with the names of all 111 cities in our two (Depth and Breadth) datasets and selected the city name with the highest similarity score based on CLIP's visual-textual embedding alignment.
The results are as shown in Table~\ref{tab:clip}.
Despite its architectural simplicity and lack of geographic priors or structured reasoning, zero-shot CLIP achieves a substantial improvement over the random baseline.
This supports the claim that vision-language alignment alone contributes meaningfully to geolocation performance.
However, CLIP still lags far behind modern VLMs (\eg, GPT-4o reaches 40–57\% accuracy on the same set), which demonstrates the necessity of more advanced multimodal reasoning and world knowledge for city-level geolocation.

\begin{table*}[h!]
\centering
\begin{tabular}{lcccc}
\toprule
\textbf{City} & \textbf{Total Score} & $d \le 100km$ & $100km < d \le 1000km$ & $d > 100km$ \\
\midrule
San Antonio    & 50 & 20 & 10 & 0  \\
San Diego      & 48 & 19 & 10 & 1  \\
Campinas       & 48 & 19 & 10 & 1  \\
Melbourne      & 41 & 20 &  1 & 9  \\
Kolkata        & 40 & 20 &  0 & 10 \\
Adelaide       & 35 & 16 &  3 & 11 \\
Chennai        & 33 & 16 &  1 & 13 \\
Yekaterinburg  & 31 & 14 &  3 & 13 \\
\bottomrule
\end{tabular}
\caption{GPT-4o scores for some selected cities.}
\label{tab:gpt_scores_examples}
\end{table*}

\subsection{Error Analysis with Distance-Based Scores}

We further illustrate the error patterns with selected cities that have similar numbers of correct predictions (\ie, $d \le 100km$), but exhibit very different types of errors, in Table~\ref{tab:gpt_scores_examples}.
These results highlight that even when models achieve similar levels of correctness at fine-grained levels (\eg, city-level hits), the types of errors vary: some are localized within-region mistakes (\eg, Campinas mispredicted as São Paulo), while others are more severe intercontinental mismatches (\eg, Melbourne predicted as a U.S. city).
This analysis complements the accuracy and entropy metrics by offering a nuanced view of model behavior and supports the need for geospatially aware evaluation metrics.

\section{Case Studies}

\subsection{Can CoT Help?}

To evaluate the performance of VLMs, we analyze their outputs using Chain-of-Thought (CoT)~\cite{kojima2022large, wei2022chain} prompts.
We present two example queries: one for Gemini and another for LLaMA.
The case study suggests that while CoT reasoning can appear logical, it is not consistently tied to the final answer.
In CoT Example (1), Gemini correctly identifies Africa’s surroundings but notes the absence of visible license plates or signs that could aid in further country or city analysis.
Despite this lack of evidence, the model still predicts the correct answer.
Conversely, in CoT Example (2), LLaMA identifies features typical of California but incorrectly predicts Santa Barbara instead of the correct answer, Los Angeles.
Across multiple examples, the elements cited in the CoT reasoning process often partially align with the final answer.
However, these elements are typically broad and fail to accurately pinpoint specific locations.
Relying solely on the reasoning process makes it challenging to determine the exact geographical location of an image.
We additionally apply direct prompting to Gemini, instructing it to identify the geographical location without invoking explicit reasoning.
Results on the breadth subset indicate that CoT prompting yields minimal performance gains, with city-level accuracy of 63.0\% using CoT and 61.0\% without it.
This suggests that the model's outputs may not stem from genuine visual reasoning but rather reflect prior knowledge of geographic patterns.

\subsection{ChatGPT-o3}

We conduct a small-scale experiment on o3 using a random sample of 10 image pairs that GPT-4o misclassifies at the city level.
The results show that o3 achieves 0\% accuracy on these images.
Notably, the misclassified images typically depict less populous or underdeveloped regions.
This suggests that o3 may exhibit a similar bias, leading to reduced accuracy for images from such areas.

\begin{table*}[h]
    \resizebox{0.48\linewidth}{!}{
    \begin{tabular}{lp{8.6cm}}
    \toprule
    \rowcolor{mygray}
    \multicolumn{2}{l}{\textbf{CoT Example (1)}} \\
    \textsc{Image} & \includegraphics[width=0.5\linewidth]{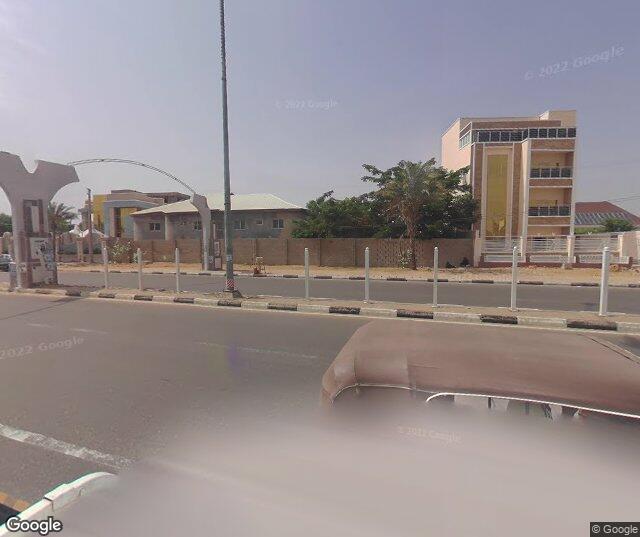} \\
    \midrule
    \textsc{Ground Truth} & Kano, Nigeria, Africa \\
    \textsc{Model} & Gemini-1.5-Pro \\
    \textsc{Analysis} & The image appears to be taken in a tropical or subtropical region due to the vegetation and sun angle. The architecture suggests a possibly African context. The road appears to be well-maintained. No license plates or signage are clearly visible to provide definitive clues. \\
    \textsc{Prediction} & Kano, Nigeria, Africa \\
    \bottomrule
    \end{tabular}
    }
    \resizebox{0.48\linewidth}{!}{
    \begin{tabular}{lp{8.6cm}}
    \toprule
    \rowcolor{mygray}
    \multicolumn{2}{l}{\textbf{CoT Example (2)}} \\
    \textsc{Image} & \includegraphics[width=0.5\linewidth]{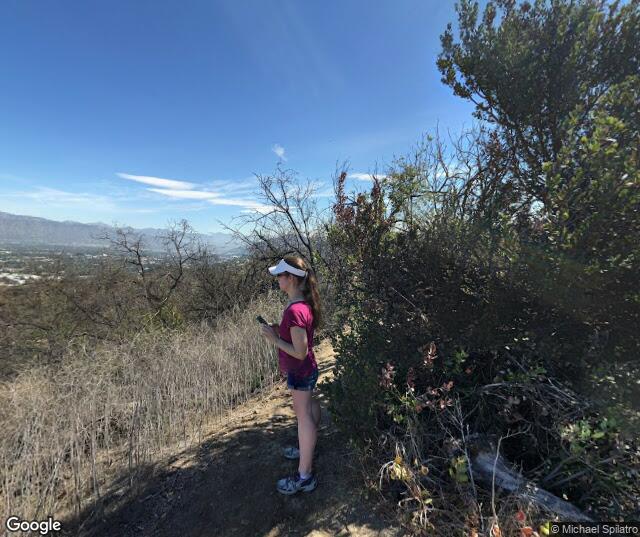} \\
    \midrule
    \textsc{Ground Truth} & Los Angeles, USA, North America \\
    \textsc{Model} & LLaMA-3.2-11B-Vision \\
    \textsc{Analysis} & The image shows a woman standing on a dirt path in a wooded area, holding a cell phone and wearing a pink t-shirt, shorts, and a white visor. She is looking out at a view of mountains and valleys in the distance. \\
    & *   The woman is wearing a pink t-shirt, shorts, and a white visor. \\
    & *   She is holding a cell phone in her right hand. \\
    & *   She is standing on a dirt path in a wooded area. \\
    & *   There are trees and bushes surrounding her. \\
    & *   In the background, there is a view of mountains and valleys. \\
    & *   The sky is blue with some clouds. \\
    & Based on the analysis, the street view appears to be located in a rural or natural area, possibly in a national park or nature reserve. The presence of mountains and valleys in the background suggests that the location is in a hilly or mountainous region. \\
    \textsc{Prediction} & Santa Barbara, USA, North America \\
    \bottomrule
    \end{tabular}
    }
\end{table*}

\begin{table*}[h]
    \resizebox{0.48\linewidth}{!}{
    \begin{tabular}{lp{8.6cm}}
    \toprule
    \rowcolor{mygray}
    \multicolumn{2}{l}{\textbf{o3 Example (1)}} \\
    \textsc{Image} & \includegraphics[width=0.5\linewidth]{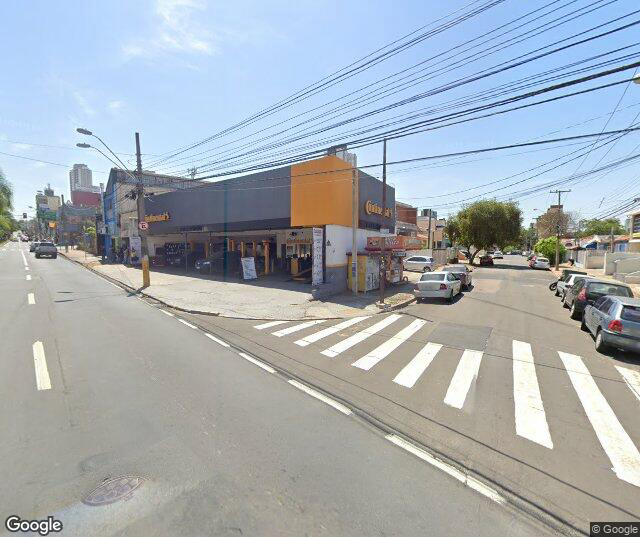} \\
    \midrule
    \textsc{Ground Truth} & Campinas, Brazil, South America \\
    \textsc{Model} & ChatGPT-o3 \\
    \textsc{Analysis} & 1) Latitude/Longitude – the sun sits high in the northern half of the sky (shadows fall slightly south-east), placing us in the southern hemisphere at roughly 22–23 ° S and 49 ° W. 2) Sun position – almost overhead, hinting the image was taken close to the Tropic of Capricorn in late morning/early afternoon. 3) Vegetation – broad-leaf street trees and palms typical of Brazil’s south-east interior; no conifers or temperate species. 4) Natural scenery – fully urban terrain on a gentle slope, no coastline or mountains in view. 5) Buildings – corner lot tyre centre painted grey/orange with the Continental logo; next door narrow two-storey shop houses; far background a 25–30-storey tower common in mid-sized Brazilian cities. 6) Licence plates – long, very narrow rectangles with a small blue Mercosur band (blurred, but shape matches current Brazilian plates). 7) Road directions – right-hand traffic, single broken centre line, Brazilian style zebra crossing and ‘E’ (Estacionar) no-parking sign. 8) Flags – none visible. 9) Language – Portuguese words on the pole (‘Proibido Estacionar’) and on the kiosk; brand signage ‘Pneus’ also Portuguese. 10) Shops – Continental / Cantu Pneus service centre; a small yellow “orelhão” pay-phone booth (a Brazilian hallmark) beside a snack kiosk. 11) Pedestrians – a few customers near the garage entrance, attire consistent with warm Brazilian climate. Combining Portuguese language, Brazilian street furniture, right-hand driving, and specific chain décor for Cantu-Continental stores, the scene fits the interior of São Paulo state.  The only Cantu-Continental branch on a left-hand corner with a narrow residential side-street and a tall tower on the same axis is on Avenida Duque de Caxias in Bauru.  That corner (with Rua Araujo Leite) matches the slope, tree and phone-booth layout seen here. \\
    \textsc{Prediction} & Bauru, Brazil, South America \\
    \bottomrule
    \end{tabular}
    }
    \resizebox{0.48\linewidth}{!}{
    \begin{tabular}{lp{8.6cm}}
    \toprule
    \rowcolor{mygray}
    \multicolumn{2}{l}{\textbf{o3 Example (2)}} \\
    \textsc{Image} & \includegraphics[width=0.5\linewidth]{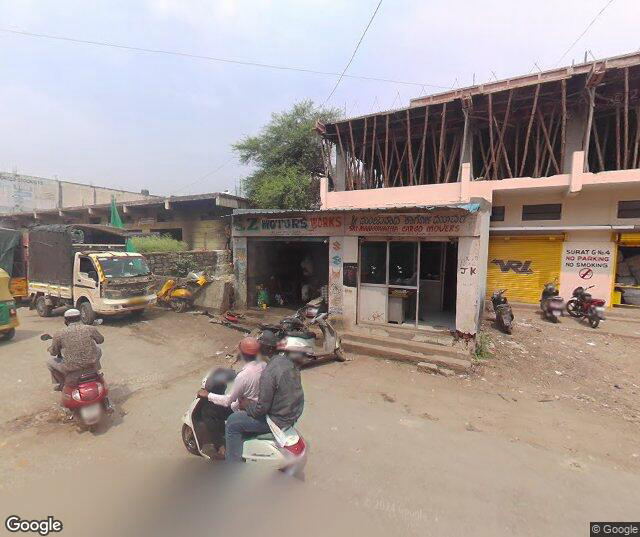} \\
    \midrule
    \textsc{Ground Truth} & Bengaluru, India, Asia \\
    \textsc{Model} & ChatGPT-o3 \\
    \textsc{Analysis} & 1) Latitude and Longitude: The signage explicitly naming Surat and the semi-arid coastal setting place the view at roughly 21.17 ° N, 72.86 ° E on the south-eastern side of the city. 2) Sun position: Short shadows that fall slightly leftward from vertical objects indicate the sun is high in the southern sky; the camera is therefore pointing roughly north-east around midday. 3) Vegetation: A single broad-leafed tree and low scrub, consistent with the dry-tropical vegetation found around coastal Gujarat. 4) Natural scenery: The terrain is flat with no hills or water visible, typical of the Surat plain near the Tapti River delta. 5) Buildings: One- and two-storey concrete workshops and godowns; one structure is under construction with exposed re-bar and wooden scaffolding—common in Indian industrial estates. 6) License plates: Vehicles show yellow (commercial) and white (private) rectangular plates in the Indian style, confirming India. ) Road directions: Traffic keeps left and vehicles are right-hand-drive, matching Indian road rules. 8) Flags: A plain green shop flag—no national flag present. 9) Language: English text ("SURAT GN-4 NO PARKING", "Z Motors Works") plus faint Gujarati script confirm the local language mix of Gujarat. 10) Shops: A two-wheeler repair garage (Z Motors Works) and a yellow-shuttered VRL Logistics cargo depot identify the spot as part of a transport yard. 11) Pedestrians: Several helmeted scooter riders and workers in typical Indian attire reinforce the urban-industrial context. Taking all of these clues together, the scene fits an interior lane (Road No 3) of the Udhna Udyog Nagar industrial estate on the southern fringe of Surat, Gujarat, India. \\
    \textsc{Prediction} & Surat, India, Asia \\
    \bottomrule
    \end{tabular}
    }
\end{table*}

\clearpage
\onecolumn

\section{User Study Questionnaire}

\begin{figure}[h!]
  \centering
  \subfloat[Instruction for human participants.]{
    \includegraphics[width=0.5\linewidth]{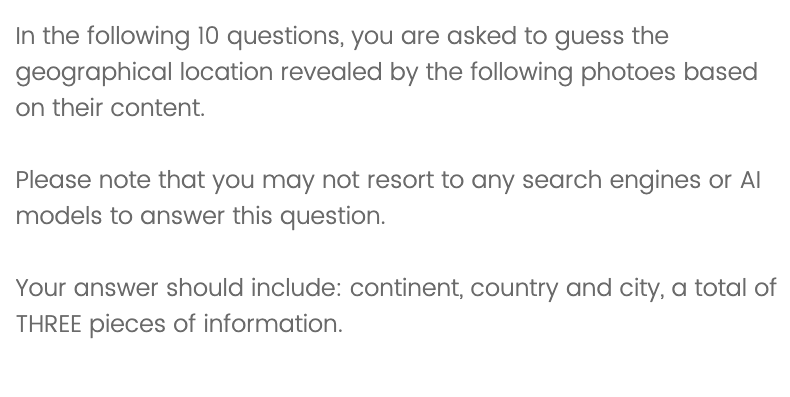}
  } \\
  \subfloat[An example question.]{
    \includegraphics[width=0.5\linewidth]{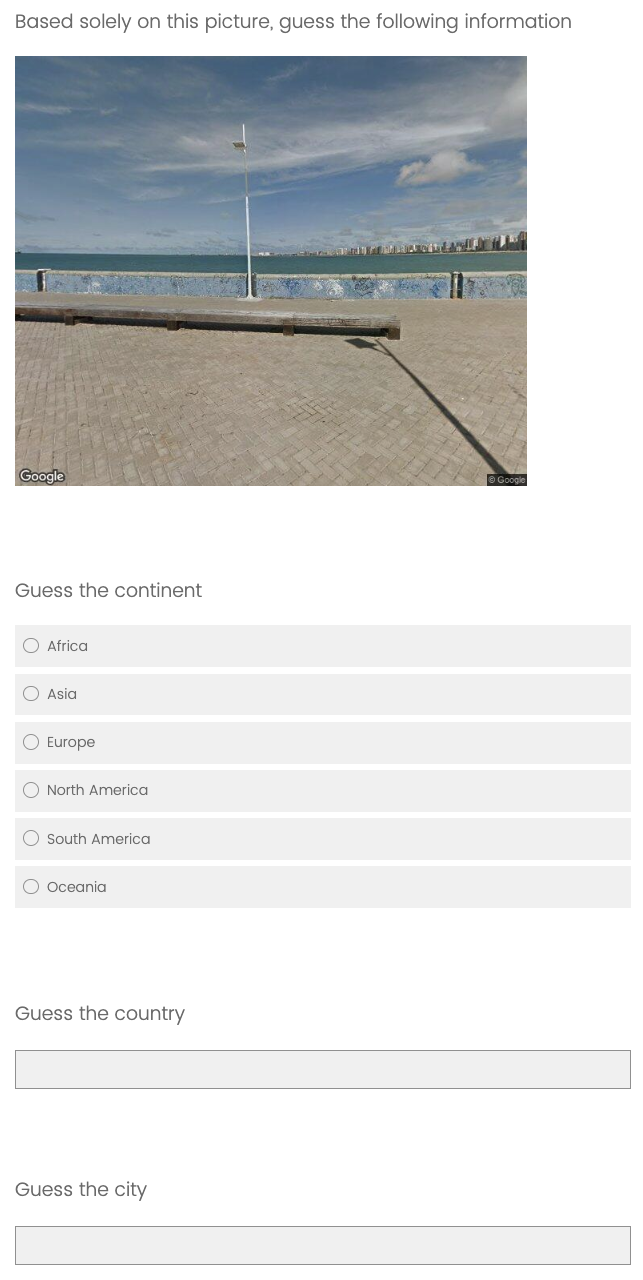}
  }
  \caption{Illustration of our questionnaires.}
  \label{fig:questionnaire}
\end{figure}

\end{document}